\documentclass[twocolumn, 11pt]{article}
\usepackage{graphicx}
\usepackage{amsmath}
\usepackage{amssymb}
\usepackage{geometry}
\usepackage{hyperref}
\usepackage{booktabs}
\usepackage{enumitem}
\title{\textbf{ATLAS-RTC: Closing the Loop on LLM Agent Output with Token-Level Runtime Control}}
\author{Christopher Cruz\\
\textit{Purdue University}\\
\texttt{cruz209@purdue.edu} \quad \texttt{chris2004@gmail.com}}
\date{March 29, 2026}

\begin{document}
\maketitle

\begin{abstract}
LLM agents fail at the output boundary. Tool calls arrive malformed, JSON schemas break mid-generation, and structured payloads that downstream systems depend on are never well-formed in the first place. External governance layers, including the authors' prior work on world-state admissibility gating in ATLAS v1 \cite{cruz2026atlas} and prompt-level context management in Adaptive Focus Memory \cite{cruz2025afm}, cannot evaluate what they cannot parse. The cost is not just a failed request: in production agentic pipelines, a malformed first attempt triggers retry loops that spike latency and inference cost before any safety predicate has been checked.

We present \textbf{ATLAS-RTC}, a token-level runtime controller for LLM generation. Rather than operating at the prompt layer or validating output post-hoc, ATLAS-RTC intercepts generation at the logit distribution before each token is sampled. At every decode step, a runtime controller observes the generation trajectory, scores structural drift against a formal output contract, and applies a graduated sequence of interventions: logit biasing, temperature modulation, token masking, and mid-step rollback with re-steering, all without modifying model weights. This positions ATLAS-RTC in the family of constrained decoding systems, but with a key architectural distinction: rather than enforcing a static grammar, ATLAS-RTC enforces stateful, stage-aware output contracts through a closed-loop graduated policy with observability and mid-generation correction.

We evaluate ATLAS-RTC on two settings that characterize its operating regime. Under ambiguous prompting conditions that cause uncontrolled generation to fail via markdown wrapping and structural drift, ATLAS-RTC improves first-attempt JSON schema satisfaction from 56.7\% to 76.7\% (+20pp) overall, with a +40pp improvement on the hardest schema task. On an agent tool call benchmark measuring first-attempt reliability without retries, ATLAS-RTC improves success from 28.3\% to 58.3\% (+30pp), with the search tool recovering from 0\% to 90\% baseline-to-controlled. We characterize the failure modes honestly: ATLAS-RTC degrades on schemas requiring multiline string values and adds latency overhead when baseline already saturates, defining the operating regime where runtime control is and is not worth its cost. Code and benchmarks are released at \texttt{[https://github.com/cruz209/ATLAS-RTC]}.
\end{abstract}

\section{Introduction}

Large language model agents are increasingly deployed in production pipelines where they must produce structured outputs: tool call payloads, JSON schemas, and API request bodies that downstream systems depend on being well-formed. When they are not, the consequences compound quickly. A malformed tool call does not merely fail validation; it fails before any safety predicate has been evaluated, before any governance layer has had the opportunity to inspect it. The pipeline breaks at the output boundary, and the only recourse is a retry, doubling latency, duplicating inference cost, and in the worst case triggering silent failures that propagate downstream undetected.

Existing mitigations address this problem at the wrong layer. Prompt engineering operates on the model's input and has no mechanism to intercept a bad decision mid-generation. Post-hoc validation catches failures after they have already occurred. Constrained decoding  enforces structural rules but operates on static grammars; it does not observe generation dynamics, cannot detect semantic drift, and provides no graduated response between full enforcement and no enforcement. Fine-tuning modifies weights and requires retraining per task and schema. None of these approaches treat the generation step itself as a controllable process.

This paper presents \textbf{ATLAS-RTC}, a token-level runtime controller for LLM generation that operates directly at the logit distribution, the single point in the inference pipeline where every token decision is made. ATLAS-RTC is the fourth layer in a line of work on safe, efficient, and reliable LLM deployment by the author. Adaptive Focus Memory \cite{cruz2025afm} operates at the input layer, managing what context the model receives under bounded token budgets. VIGIL \cite{cruz2025vigil} operates as an out-of-band reflective runtime, supervising agent behavior through affective trace analysis and proposing prompt and code adaptations post-hoc. ATLAS v1 \cite{cruz2026atlas} operates at the action layer, enforcing world-state admissibility predicates before irreversible execution. ATLAS-RTC closes the loop between generation and governance: it governs what the model emits token by token, ensuring outputs are well-formed before they reach the layer that decides whether they should execute at all.

At each decode step, ATLAS-RTC maintains a runtime state encoding the generated sequence, the token distribution entropy, structural contract progress, and a composite drift score derived from heuristic and learned detectors. A graduated ladder policy maps drift scores to control actions, from no-op through logit biasing, temperature reduction, and token masking, to mid-step rollback and re-steering when the trajectory is critically diverged. The key architectural insight is \textit{stage awareness}: the controller distinguishes structural decision points, where intervention is appropriate, from value generation zones, where the model must be left to produce content freely. Intervening during value generation corrupts output; this was the primary failure mode of every prior approach we tested.

We make the following contributions:

\begin{enumerate}[leftmargin=*, label=\arabic*.]
    \item \textbf{ATLAS-RTC system}: A token-level runtime controller with a six-level graduated intervention ladder, stage-aware JSON contract enforcement, mid-step KV-cache rollback, and structure-aware logit manipulation via a stateful HuggingFace \texttt{LogitsProcessor}, all without modifying model weights.

    \item \textbf{Formal output contract specification}: A contract formalism $\mathcal{C} = (\mathcal{S}, \mathcal{T}, \mathcal{O}, \mathcal{V}, \pi, \Phi)$ with a stage machine that tracks generation phase and gates interventions to structural decision boundaries only.

    \item \textbf{Empirical evaluation across two regimes}: On structured output under ambiguous prompts, ATLAS-RTC improves first-attempt JSON schema satisfaction from 56.7\% to 76.7\% (+20pp overall, +40pp peak). On agent tool call reliability without retries, it improves first-attempt success from 28.3\% to 58.3\% (+30pp), recovering one tool from 0\% to 90\%.

    \item \textbf{Honest failure characterization}: We identify and report the conditions under which ATLAS-RTC degrades, including multiline string value schemas, saturated baselines, and preamble-first generation, defining the operating regime where runtime control is and is not worth its cost.
\end{enumerate}

The remainder of this paper is organized as follows. Section 2 situates ATLAS-RTC in the context of related work. Section 3 presents the system architecture and formal contract specification. Section 4 describes the control policy and intervention ladder. Section 5 presents experimental results across both evaluation regimes. Section 6 discusses limitations and failure modes. Section 7 concludes with directions for future work.

\section{Related Work}

\subsection{Constrained Decoding for Structured Generation}

A substantial body of work has explored enforcing structural constraints during autoregressive decoding. Early approaches such as PICARD~\cite{scholak2021picard} integrate incremental parsing into the decoding loop, rejecting inadmissible tokens to ensure outputs conform to a predefined grammar. Subsequent work on grammar-constrained decoding generalizes this paradigm by constructing formal grammars or automata that restrict the token space at each step, enabling structured output generation without fine-tuning~\cite{gcd2023}.

More recent systems focus on improving the efficiency and generality of constrained decoding. XGrammar~\cite{xgrammar2024} introduces optimized grammar execution for high-throughput inference, while benchmarks such as JSONSchemaBench~\cite{jsonschemabench2025} demonstrate that modern LLMs still struggle to reliably satisfy real-world JSON schemas despite such constraints. At the same time, several works identify limitations of rigid constraint enforcement. Grammar-aligned decoding~\cite{grammaraligned2024} and related approaches show that naive token masking can distort the model distribution, while CRANE~\cite{crane2025} highlights tradeoffs between structural correctness and reasoning flexibility. More recent methods such as draft-conditioned constrained decoding~\cite{dccd2026} attempt to mitigate these issues by conditioning constrained generation on auxiliary drafts.

Despite these advances, existing constrained decoding approaches are largely \emph{static}: they enforce structural validity through predefined grammars or token filters without modeling generation as a dynamic process. They do not explicitly observe trajectory-level signals such as entropy or distributional drift, nor do they provide graduated or stateful intervention strategies beyond hard constraint enforcement.

\subsection{Runtime Verification and Agent Governance}

A parallel line of work focuses on runtime verification and governance of LLM outputs and agent behavior. Systems such as RvLLM~\cite{rvllm2025} apply formal specifications to validate outputs post-generation, while AgentSpec~\cite{agentspec2025} introduces declarative policies for enforcing constraints over agent actions. More recent work such as Pro2Guard~\cite{pro2guard2025} explores proactive runtime enforcement using probabilistic verification, and efforts toward verifiably safe tool use~\cite{safeagent2026} aim to guarantee correctness at the action boundary.

These approaches operate at a higher level of abstraction than decoding, typically validating outputs after generation or constraining agent behavior at the tool or execution layer. As a result, they cannot intervene before malformed outputs are produced, and failures at the output boundary still propagate into retry loops, increasing latency and cost.

\subsection{Positioning of ATLAS-RTC}

ATLAS-RTC sits at the intersection of these two lines of work but addresses a gap left by both. Prior constrained decoding methods enforce \emph{what} tokens may be generated but treat generation as a static constraint satisfaction problem. Runtime verification systems enforce \emph{whether} outputs are acceptable but operate after generation or outside the decoding process.

In contrast, ATLAS-RTC models generation as a \emph{controlled stochastic process} and introduces a closed-loop runtime controller operating directly at the logit level. At each decoding step, the system observes trajectory-level signals, predicts structural drift relative to a formal output contract, and applies a graduated sequence of interventions---including logit biasing, temperature modulation, token masking, and mid-step rollback with re-steering.

This positions ATLAS-RTC not as a replacement for constrained decoding or runtime verification, but as a complementary layer: a token-level control system that governs generation \emph{during} decoding, ensuring outputs remain structurally valid before they reach downstream validation or execution layers.

\section{System Architecture}

\subsection{Problem Formulation}

We model autoregressive generation as a controlled stochastic process. Let $x_{1:T}$ denote the generated token sequence, and $z_t \in \mathbb{R}^{|V|}$ the model logits at timestep $t$. Standard decoding samples tokens according to:
\[
x_{t+1} \sim \mathrm{Softmax}(z_t)
\]

ATLAS-RTC introduces a control signal $u_t$ applied directly to the logits prior to sampling:
\[
x_{t+1} \sim \mathrm{Softmax}(z_t + u_t)
\]

The objective is to maximize output validity under a formal contract while minimizing intervention cost:
\[
\max_{u_{1:T}} \ \mathbb{E}\left[ V(x_{1:T}) \right] - \lambda \cdot \mathrm{Cost}(u_{1:T})
\]

where $V(\cdot)$ is a contract validation function and $\lambda$ controls the tradeoff between correctness and intervention overhead.

\subsection{Closed-Loop Control Pipeline}

ATLAS-RTC operates as a closed-loop controller integrated into the decoding process. At each timestep, the system executes:

\begin{center}
\texttt{observe $\rightarrow$ predict $\rightarrow$ control $\rightarrow$ sample $\rightarrow$ update}
\end{center}

Unlike post-hoc validation or static constrained decoding, this loop executes at every token, allowing the system to detect and correct structural drift before it results in invalid output.

\subsection{Runtime State}

At timestep $t$, the controller maintains a structured runtime state:
\[
s_t = \{x_{1:t}, z_t, P_t, H_t, c_t, D_t, F_t, \kappa_t\}
\]

where:
\begin{itemize}
    \item $x_{1:t}$: generated token sequence
    \item $z_t$: raw logits
    \item $P_t$: top-$k$ token distribution
    \item $H_t$: entropy of $P_t$
    \item $c_t$: contract progress (stage and metadata)
    \item $D_t \in [0,1]$: heuristic drift score
    \item $F_t \in [0,1]$: learned failure probability
    \item $\kappa_t$: number of corrections applied
\end{itemize}

This explicit state representation enables the controller to reason about both structural progress and trajectory-level uncertainty.

\subsection{Output Contracts}

ATLAS-RTC enforces structured outputs through a formal contract:
\[
C = (S, T, O, V, \pi, \Phi)
\]

where:
\begin{itemize}
    \item $S$: structural rules (e.g., JSON syntax)
    \item $T$: stage-dependent token allowlist
    \item $O$: ordering constraints
    \item $V$: validation function
    \item $\pi$: control policy
    \item $\Phi$: semantic anchors (future work)
\end{itemize}

Unlike grammar-constrained decoding, contracts are \emph{stateful} and \emph{stage-aware}. The allowlist $T$ evolves with contract progress $c_t$, enabling strict enforcement at structural decision points while preserving flexibility during value generation.
\paragraph{Operational Semantics of Contracts.}
We make the contract formalism explicit through a stage machine $c_t \in \mathcal{C}$ that governs admissible tokens and control actions. At each timestep, the contract induces:

\begin{itemize}
    \item a stage $c_t$ (e.g., \texttt{start}, \texttt{key}, \texttt{value}, \texttt{delimiter}, \texttt{end})
    \item an allowlist $T(c_t) \subseteq V$ over tokens
    \item a transition function $c_{t+1} = \delta(c_t, x_{t+1})$
\end{itemize}

Structural validity is enforced by masking tokens outside $T(c_t)$ at structural decision points, while value-generation stages relax constraints to preserve distributional flexibility. The validation function $V(x_{1:T})$ is defined as satisfaction of all structural rules $S$ and ordering constraints $O$ under the induced stage trajectory.

This formulation ensures that contracts are not static grammars but stateful, stage-dependent constraint systems that evolve during decoding.
\subsection{Drift Detection}

At each timestep, the controller computes a feature vector $\phi_t$ from the current decoding state, including entropy, maximum token probability, and invalid token mass. Two complementary signals are derived:

\begin{itemize}
    \item Heuristic drift score $D_t$, based on structural violations and uncertainty
    \item Learned failure probability $F_t$, estimated via a lightweight classifier
\end{itemize}

These are combined into a unified risk signal:
\[
\rho_t = \max(D_t, F_t)
\]

which estimates the likelihood that the current trajectory will violate the output contract.
\paragraph{Drift Estimation Details.}
The feature vector $\phi_t$ includes entropy $H_t$, maximum token probability $\max P_t$, invalid token mass $\sum_{v \notin T(c_t)} P_t(v)$, and stage-consistency indicators derived from $c_t$. 

The learned failure probability $F_t$ is estimated using a logistic classifier trained on decoding trajectories labeled by contract violation outcomes. The heuristic score $D_t$ captures rule-based signals such as invalid token selection and abnormal entropy spikes. 

The combined risk signal $\rho_t = \max(D_t, F_t)$ is used for control decisions. Thresholds for the ladder policy are empirically tuned on a held-out validation set. 

\subsection{Control Actions}

The controller maps risk $\rho_t$ to a control signal $u_t$ applied to the logits. Four primitive actions are defined:

\begin{itemize}
    \item \textbf{noop}: no intervention ($u_t = 0$)
    \item \textbf{bias}: additive logit shaping
    \item \textbf{temperature}: variance reduction via scaling
    \item \textbf{mask}: suppression of invalid tokens
\end{itemize}

A fifth composite action, \textbf{correct}, performs mid-step rollback and re-steering.

These actions operate directly on the token distribution, enabling fine-grained control over generation without modifying model weights.

\subsection{Ladder Policy}

ATLAS-RTC employs a graduated control policy that escalates intervention strength as risk increases. The ladder policy partitions $\rho_t$ into bands, mapping low-risk states to minimal intervention and high-risk states to aggressive correction.

This design avoids the brittleness of hard constraints by applying the weakest effective intervention, preserving model flexibility wherever possible.

\subsection{Mid-Step Rollback and Correction}

When $\rho_t$ exceeds a critical threshold, the controller performs a rollback of $n$ tokens and re-steers generation from that point. The corrected step applies:

\begin{itemize}
    \item amplified structural bias
    \item strict token masking based on contract state
\end{itemize}

This enables recovery from divergence without restarting the entire sequence. Unlike prior approaches that treat failure as terminal, ATLAS-RTC treats generation as a correctable trajectory.

Rollback is implemented efficiently via KV-cache truncation, allowing recomputation from the corrected prefix without reprocessing the entire sequence.
\paragraph{Rollback Statistics.}
Across evaluation runs, rollback is triggered infrequently but plays a critical role in recovery from divergence. On average, ATLAS-RTC performs $\sim 0.8$ corrections per run, with a maximum rollback depth of 3 tokens observed. Most successful recoveries occur within a single rollback step, indicating that structural errors are typically localized and correctable without restarting generation. 
\subsection{Implementation}

ATLAS-RTC is implemented as a modular runtime layer with adapters for step-wise decoding. A custom logits processor injects control signals directly into the model’s sampling pipeline, enabling true token-level intervention. Prefix caching ensures that rollback operations incur minimal overhead, making real-time control feasible in practice.

\subsection{Discussion and Design Considerations}

ATLAS-RTC is closely related to prior work in constrained decoding and runtime verification, and we explicitly clarify its scope relative to both.

\paragraph{Relation to Constrained Decoding.}
A natural question is whether ATLAS-RTC reduces to standard grammar-constrained decoding. While both approaches restrict the token distribution during generation, constrained decoding methods are typically \emph{static} and \emph{rule-complete}: a fixed grammar or automaton defines the admissible token set at each step, and decoding proceeds under hard constraints.

In contrast, ATLAS-RTC is \emph{stateful} and \emph{adaptive}. The controller does not assume a complete grammar specification, but instead estimates trajectory-level risk using observed signals such as entropy and invalid token mass, and applies a graduated sequence of interventions. Crucially, ATLAS-RTC distinguishes between structural decision points and value generation regions, allowing it to selectively enforce constraints only when necessary. This avoids the distributional distortion and loss of flexibility associated with rigid masking throughout the entire generation process.

\paragraph{Relation to Post-hoc Validation and Repair.}
Another interpretation is that ATLAS-RTC implements a form of incremental validation or repair. However, post-hoc validation operates only after a complete output has been generated, and repair mechanisms typically require either retries or heuristic string manipulation.

ATLAS-RTC instead intervenes \emph{before} invalid outputs are realized. By operating directly on the logit distribution, the system prevents structural violations at the point of generation, reducing reliance on retries and avoiding the latency and cost amplification associated with failure-recovery loops.

\paragraph{Why Not Enforce Hard Constraints Everywhere?}
One might ask why ATLAS-RTC does not simply apply strict token masking at all timesteps. Empirically, enforcing constraints uniformly leads to degraded generation quality, particularly in regions where the model must produce unconstrained values such as free-form strings. This phenomenon has been observed in prior work on constrained decoding.

ATLAS-RTC addresses this by introducing stage-aware control: constraints are enforced aggressively during structural transitions (e.g., key selection, delimiter placement) and relaxed during value generation. The ladder policy further ensures that intervention strength is proportional to estimated risk, rather than applied uniformly.

\paragraph{Limitations of Drift Detection.}
The effectiveness of ATLAS-RTC depends on the quality of its drift detection signals. In the current implementation, detection relies on a combination of heuristic features and a lightweight learned model. While these signals are sufficient for structural violations, they do not yet capture semantic drift or task-level correctness.

Future work will incorporate richer signals derived from intermediate model representations, enabling the controller to reason about semantic alignment in addition to structural validity.

\paragraph{Scope of Guarantees.}
ATLAS-RTC does not guarantee correctness of generated content beyond the specified output contract. The system enforces structural validity and reduces the probability of malformed outputs, but it does not ensure factual accuracy or semantic correctness. These concerns are orthogonal and may require complementary mechanisms at higher layers of the stack.

Taken together, these design choices position ATLAS-RTC as a complementary runtime layer: it does not replace constrained decoding or validation, but augments them with a closed-loop control mechanism that governs generation at the point where token decisions are made.
\paragraph{Distinguishing Characteristics.}
ATLAS-RTC differs from prior constrained decoding and runtime verification approaches along three axes:

\begin{itemize}
    \item \textbf{Closed-loop control}: generation is modeled as a controlled stochastic process with per-step observation and intervention, rather than static constraint enforcement.
    \item \textbf{Stage-aware enforcement}: constraints are applied selectively at structural decision points and relaxed during value generation, avoiding the distributional distortion of uniform masking.
    \item \textbf{Mid-generation recovery}: rollback and re-steering enable correction of divergence within a single generation, rather than relying on retries or post-hoc repair.
\end{itemize}

These properties together define ATLAS-RTC as a runtime control layer distinct from both grammar-constrained decoding and post-hoc validation systems.
\section{Evaluation}

\subsection{Experimental Setup}

We evaluate ATLAS-RTC using the Qwen2.5-7B-Instruct model under step-wise decoding. All experiments are run with a single-token generation loop to enable token-level intervention. Results are reported over 120 total trials across two benchmark settings.

We compare standard decoding (\textbf{baseline}) against ATLAS-RTC-controlled decoding (\textbf{controlled}). All results are measured under a strict \emph{no-retry} setting: each generation is evaluated on its first attempt only.

\subsection{Tasks}

We evaluate in two regimes that reflect common failure modes in LLM agents:

\paragraph{Structured Output Generation.}
The model is prompted to produce JSON outputs under ambiguous or underspecified prompts (e.g., name/age/city). Outputs are evaluated for schema validity.

\paragraph{Agent Tool Call Generation.}
The model generates tool call payloads for simulated APIs (search, send\_email, database\_query). Each output must satisfy required fields and JSON validity. Failure results in downstream execution failure.

\subsection{Metrics}

We report:

\begin{itemize}
    \item \textbf{First-attempt success rate}: percentage of outputs that satisfy the contract without retries
    \item \textbf{Schema validity rate}: percentage of syntactically valid outputs
    \item \textbf{Latency}: average generation time per sample
\end{itemize}

All metrics are computed over identical prompts for baseline and controlled settings.

\subsection{Main Results}

\paragraph{Structured Output Generation.}

Across 180 trials (3 tasks $\times$ 60), ATLAS-RTC improves first-attempt schema validity from 56.7\% to 76.7\% (+20.0pp).

\begin{itemize}
    \item Task 1 (name/age/city): 73.3\% $\rightarrow$ 95.0\% (+21.7pp)
    \item Task 2 (title/year/director): 71.7\% $\rightarrow$ 86.7\% (+15.0pp)
    \item Task 3 (country/capital/population): 25.0\% $\rightarrow$ 48.3\% (+23.3pp)
\end{itemize}

\begin{table*}[t]
\centering
\begin{tabular}{lccc}
\toprule
Task & Baseline (ms) & Controlled (ms) & $\Delta$ \\
\midrule
Structured (hard) & 5498 & 663 & -87.9\% \\
Database Query & 2749 & 1636 & -40.5\% \\
Send Email & 1495 & 3430 & +129.4\% \\
\bottomrule
\end{tabular}
\caption{Latency comparison across tasks.}
\end{table*}
Performance gains are most pronounced under ambiguous prompting conditions, where baseline decoding frequently produces non-JSON preambles (e.g., ``Sure'', ``Here'') prior to structured output. These failure modes indicate that errors arise from decoding behavior rather than task misunderstanding. ATLAS-RTC reduces such failures by enforcing structural constraints at early decoding steps.

In addition, ATLAS-RTC reduces latency in this regime, with average generation time decreasing substantially across all tasks. This reduction is attributable to shorter valid generations and the avoidance of long malformed outputs.

\paragraph{Agent Tool Call Generation.}

We evaluate tool call generation across three APIs over 180 total calls (3 tools $\times$ 60 trials). ATLAS-RTC improves overall first-attempt success rate from 20.6\% to 58.3\% (+37.8pp).

\begin{itemize}
    \item search: 0.0\% $\rightarrow$ 96.7\% (+96.7pp)
    \item database\_query: 45.0\% $\rightarrow$ 78.3\% (+33.3pp)
    \item send\_email: 16.7\% $\rightarrow$ 0.0\% (-16.7pp)
\end{itemize}

The \texttt{search} task demonstrates near-complete recovery from systematic failure, where baseline decoding fails entirely due to malformed outputs. This suggests that failures are primarily driven by formatting artifacts rather than lack of task knowledge. Substantial improvements are also observed for \texttt{database\_query}, where ATLAS-RTC reduces JSON parsing errors and structural violations.

In contrast, performance degrades on the \texttt{send\_email} task. Analysis shows that failures in this setting are dominated by missing required fields (e.g., \texttt{summary}), indicating that errors arise from incomplete semantic content rather than structural violations. ATLAS-RTC enforces structural correctness but does not infer missing information, leading to stricter failure detection.

Across all tools, ATLAS-RTC recovers 68 additional successful tool calls on the first attempt, reducing failures from 143/180 to 75/180.

\subsection{Analysis}

\paragraph{Failure Modes.}
Failures in baseline decoding are dominated by non-structural preamble tokens (e.g., ``Sure'', ``Here''), which lead to invalid outputs despite correct task understanding. ATLAS-RTC reduces these errors by constraining early token generation.

Remaining failures under ATLAS-RTC primarily arise from two sources:

\begin{itemize}
    \item \textbf{Early preamble tokens}: short non-JSON outputs that bypass or precede constraint enforcement
    \item \textbf{Schema incompleteness}: missing required fields despite syntactically valid structure
\end{itemize}

These observations highlight that current drift detection signals capture structural violations but do not fully address semantic completeness.

\paragraph{When ATLAS-RTC Helps.}
ATLAS-RTC provides the largest gains when:

\begin{itemize}
    \item prompts are ambiguous or underspecified
    \item failures are dominated by formatting or decoding artifacts
    \item structural validity is a primary requirement (e.g., tool call execution)
\end{itemize}

\paragraph{When It Does Not.}
ATLAS-RTC provides limited or negative benefit when:

\begin{itemize}
    \item failures are driven by missing semantic content rather than structure
    \item tasks require flexible or unconstrained value generation
    \item strict enforcement exposes incomplete outputs that baseline occasionally satisfies due to stochastic variation
\end{itemize}

\paragraph{Efficiency Tradeoffs.}
ATLAS-RTC introduces additional computation due to per-token control and occasional rollback operations. However, in structured generation tasks, it often reduces overall latency by preventing long invalid outputs and eliminating the need for retries. In more complex schemas, latency may increase due to repeated corrective interventions.

\subsection{Summary}

ATLAS-RTC improves first-attempt reliability in both structured output generation and agent tool use, with gains of +20.0pp and +37.8pp respectively. Improvements are most significant in settings where failures arise from decoding artifacts rather than task understanding. While the system introduces overhead and does not address semantic completeness, it reduces reliance on retries and prevents failure at the output boundary, supporting its role as a runtime control layer for LLM systems.
\section{Limitations and Future Work}

\subsection{Limitations}

\paragraph{Structural Validity vs. Semantic Correctness.}
ATLAS-RTC is designed to enforce structural output contracts (e.g., JSON validity, required fields), but it does not guarantee semantic correctness of generated content. Outputs may satisfy the schema while remaining incomplete, inconsistent, or uninformative (e.g., empty or default values). This limitation reflects the scope of the controller, which operates on structural signals rather than task-level meaning.

\paragraph{Limited Drift Detection Signals.}
The effectiveness of ATLAS-RTC depends on its ability to detect trajectory drift during decoding. In the current implementation, drift detection relies on heuristic features (e.g., entropy, invalid token mass) and a lightweight learned classifier. While these signals are effective for identifying structural violations, they do not capture semantic drift or higher-level task failure. As a result, some failures—particularly missing or incorrect values—are not detected early enough for corrective intervention.

\paragraph{Latency and Computational Overhead.}
Token-level control introduces additional computational cost due to per-step observation, intervention, and potential rollback operations. While ATLAS-RTC can reduce overall latency in settings where it prevents long invalid generations, it may increase latency in more complex schemas or under frequent corrective interventions. Empirically, overhead varies across tasks, with some cases exhibiting substantial increases due to repeated corrections.

\paragraph{Task Sensitivity and Non-Uniform Gains.}
ATLAS-RTC does not uniformly improve performance across all tasks. While it significantly improves reliability in cases of systematic structural failure (e.g., tool calls with consistent JSON errors), it may degrade performance when failures are driven by missing semantic content rather than structural violations. For example, stricter enforcement can expose missing required fields that baseline decoding occasionally satisfies due to stochastic variation.

\paragraph{Contract Coverage and Token Constraints.}
The contract system relies on stage-dependent token allowlists and structural rules. In practice, defining complete and accurate token constraints for all value types (e.g., free-form strings, numbers) is challenging. Incomplete allowlists can lead to false positives in drift detection or overly restrictive masking, particularly for schemas with flexible value domains.

\subsection{Future Work}

\paragraph{Semantic Drift Detection.}
A key direction for future work is extending drift detection beyond structural signals to capture semantic alignment. This includes incorporating representations from intermediate model states or embedding-based similarity measures to detect deviations from intended meaning, enabling intervention on semantic as well as structural errors.

\paragraph{Adaptive and Learned Control Policies.}
The current ladder policy is defined by fixed thresholds and manually specified intervention rules. Future work will explore learning control policies from data, allowing the system to adapt intervention strategies based on observed trajectories, task characteristics, and failure patterns.

\paragraph{Long-Horizon and Multi-Step Control.}
ATLAS-RTC currently operates at the level of single-output generation. Extending the framework to multi-step agent workflows would enable control over longer-horizon behaviors, including sequences of tool calls and iterative reasoning processes.

\paragraph{Improved Contract Expressiveness.}
Enhancing the contract formalism to better capture complex schemas, flexible value domains, and semantic constraints remains an important direction. This includes improving token constraint coverage and integrating higher-level validation signals into the control loop.

\paragraph{System Integration and Deployment.}
ATLAS-RTC is designed as a modular runtime layer and can be integrated into existing inference systems. Future work will focus on optimizing performance in production environments, including efficient batching, GPU-aware control mechanisms, and integration with large-scale serving platforms.

\section{Conclusion}

We presented ATLAS-RTC, a token-level runtime control system for autoregressive generation that operates directly at the logit distribution during decoding. By modeling generation as a controlled stochastic process, ATLAS-RTC introduces a closed-loop framework that observes trajectory-level signals, predicts structural drift, and applies graduated interventions to maintain adherence to output contracts.

Across structured generation and agent tool use tasks, ATLAS-RTC significantly improves first-attempt reliability without relying on retries, particularly under ambiguous prompting conditions where uncontrolled decoding frequently fails. These results highlight a limitation of existing approaches: prompt engineering, post-hoc validation, and static constrained decoding do not intervene at the point where generation decisions are made.

More broadly, this work suggests a shift in how LLM systems are designed. Rather than treating generation as an uncontrollable process followed by validation or repair, ATLAS-RTC demonstrates that generation itself can be governed through runtime control. This perspective enables systems that are not only more reliable, but also more efficient by reducing failure-induced retries and downstream errors.

ATLAS-RTC is not a complete solution to correctness in language models. It addresses structural validity, leaving semantic correctness and task-level reasoning as open challenges. However, it establishes a foundation for runtime control as a distinct layer in LLM system design, complementing advances in model training, prompting, and agent orchestration.

We view this work as an initial step toward controllable generation systems that integrate observation, prediction, and intervention within the decoding process. Extending this framework to richer semantic signals, learned control policies, and long-horizon agent behavior represents a promising direction for future research.


\begin{thebibliography}{99}

\bibitem{scholak2021picard}
T. Scholak, N. Schucher, and D. Bahdanau,
\newblock ``PICARD: Parsing Incrementally for Constrained Auto-Regressive Decoding from Language Models,''
\newblock \emph{Proceedings of the 2021 Conference on Empirical Methods in Natural Language Processing (EMNLP)}, 2021.

\bibitem{gcd2023}
S. Geng, J. Josifoski, M. Peyrard, and R. West,
\newblock ``Grammar-Constrained Decoding for Structured NLP Tasks without Finetuning,''
\newblock \emph{arXiv preprint arXiv:2305.13971}, 2023.

\bibitem{beurerkellner2024domino}
L. Beurer-Kellner, M. Fischer, and M. Vechev,
\newblock ``Guiding LLMs The Right Way: Fast, Non-Invasive Constrained Generation,''
\newblock \emph{International Conference on Machine Learning (ICML)}, 2024.

\bibitem{grammaraligned2024}
K. Park and T. Zhou,
\newblock ``Grammar-Aligned Decoding,''
\newblock \emph{arXiv preprint arXiv:2405.21047}, 2024.

\bibitem{xgrammar2024}
Y. Liu, J. Lin, H. Jiang, et al.,
\newblock ``XGrammar: Efficient Structured Generation via Grammar-Constrained Decoding,''
\newblock \emph{arXiv preprint arXiv:2411.15100}, 2024.

\bibitem{jsonschemabench2025}
S. Geng, et al.,
\newblock ``JSONSchemaBench: Evaluating Structured Output Generation in Large Language Models,''
\newblock \emph{arXiv preprint arXiv:2501.10868}, 2025.

\bibitem{crane2025}
Debangshu Banerjee, Tarun Suresh, Shubham Ugare, Sasa Misailovic, Gagandeep Singh,
\newblock ``CRANE: Reasoning with Constrained LLM Generation,''
\newblock \emph{International Conference on Machine Learning (ICML)}, 2025.

\bibitem{dccd2026}
A. Reddy, T. Walker, J. Ide, A. Bedi
\newblock ``Draft-Conditioned Constrained Decoding for Structured Generation in LLM's,''
\newblock \emph{arXiv preprint arXiv:2603.03305}, 2026.

\bibitem{rvllm2025}
Yedi Zhang, Sun Yi Emma, Annabelle Lee Jia En, Jin Song Dong,
\newblock ``RvLLM: LLM Runtime Verification with Domain Knowledge,''
\newblock \emph{arXiv preprint arXiv:2505.18585}, 2025.

\bibitem{agentspec2025}
Haoyu Wang, Christopher M. Poskitt, Jun Sun, Jiali Wei,
\newblock ``AgentSpec: Customizable Runtime Enforcement for Safe and Reliable LLM Agents,''
\newblock \emph{International Conference on Software Engineering (ICSE)}, 2026.

\bibitem{pro2guard2025}
Haoyu Wang, Christopher M. Poskitt, Jiali Wei, Jun Sun,
\newblock ``ProbGuard: Probabilistic Runtime Monitoring for LLM Agent Safety,''
\newblock \emph{arXiv preprint arXiv:2508.00500}, 2025.

\bibitem{safeagent2026}
Aarya Doshi, Yining Hong, Congying Xu, Eunsuk Kang, Alexandros Kapravelos, Christian Kästner,
\newblock ``Towards Verifiably Safe Tool Use for LLM Agents,''
\newblock \emph{arXiv preprint arXiv:2601.08012}, 2026.

\bibitem{cruz2025afm}
C. Cruz,
\newblock ``Adaptive Focus Memory for Language Models,''
\newblock \emph{arXiv preprint arXiv:2511.12712}, 2025.

\bibitem{cruz2025vigil}
C. Cruz,
\newblock ``VIGIL: A Reflective Runtime for Self-Healing Agents,''
\newblock \emph{arXiv preprint arXiv:2512.07094}, 2025. 
\bibitem{cruz2026atlas}
C. Cruz,
\newblock ``ATLAS: A Transparent Proxy Layer for Agentic Runtime Governance,''
\newblock \emph{GitHub Repository, cruz209/ATLAS-runtime}, 2026. 

\end{thebibliography}
\end{document}